\documentclass[journal]{IEEEtran}

\usepackage{graphicx}
\usepackage{stfloats}
\usepackage{booktabs}
\usepackage{multirow}
\usepackage{bbding}
\usepackage{amsmath}
\usepackage{xcolor}
\usepackage[numbers,sort&compress]{natbib}
\usepackage[colorlinks=true, citecolor=black, linkcolor = black, anchorcolor = black, urlcolor=darkgray]{hyperref}

\ifCLASSINFOpdf
\else
\fi

\begin{document}
\title{Towards Contextual Spelling Correction for Customization of End-to-end Speech \\Recognition Systems}
\author{Xiaoqiang~Wang, Yanqing~Liu, Jinyu~Li, Veljko~Miljanic, Sheng~Zhao, Hosam~Khalil
\thanks{Xiaoqiang Wang, Yanqing Liu, and Sheng Zhao are with Microsoft, China (e-mail: \{xiaoqwa, yanqliu, szhao\}@microsoft.com).}
\thanks{Jinyu Li, Veljko Miljanic and Hosam Khalil are with Microsoft, US (e-mail: \{jinyli, veljkom, hosamk\}@microsoft.com).}
\thanks{Manuscript received xxx xxx, xxx; revised xxx xxx, xxx.}}


\maketitle

\begin{abstract}
Contextual biasing is an important and challenging task for end-to-end automatic speech recognition (ASR) systems, which aims to achieve better recognition performance by biasing the ASR system to particular context phrases such as person names, music list, proper nouns, etc.
Existing methods mainly include contextual LM biasing and adding bias encoder into end-to-end ASR models.
In this work, we introduce a novel approach to do contextual biasing by adding a contextual spelling correction model on top of the end-to-end ASR system. 
We incorporate contextual information into a sequence-to-sequence spelling correction model with a shared context encoder. The proposed model includes two different mechanisms: autoregressive (AR) and non-autoregressive (NAR). We also propose filtering algorithms to handle large-size context lists, and performance balancing mechanisms to control the biasing degree of the model. The proposed model is a general biasing solution which is domain-insensitive and can be adopted in different scenarios.
Experiments show that the proposed method achieves as much as 51\% relative word error rate (WER) reduction over ASR system and outperforms traditional biasing methods. Compared to the AR solution, the NAR model reduces model size by 43.2\% and speeds up inference by 2.1 times.
\end{abstract}


\IEEEpeerreviewmaketitle

\section{Introduction}

\IEEEPARstart{I}{n} recent years, end-to-end (E2E) ASR systems \cite{li2021recent} have obtained significant improvements and achieved performance comparable to traditional hybrid systems \cite{sainath2020streaming, SC_4}. Some representative works include Attention-based Encoder-Decoder (AED) \cite{cho2014learning, Attention-bahdanau2014, LAS}, recurrent neural network Transducer (RNN-T) \cite{OTF_rescore3_0, Li2019RNNT, saon2020alignment, zeyer2020new} and transformer transducer (T-T) \cite{yeh2019transformer, zhang2020transformer, chen2020developing}. 
However, it's still challenging for E2E models to incorporate contextual information which is dynamic and domain related. Such information may be person names in a personal assistant, commonly used terms in a specific domain, proper nouns and so on. 
E2E ASR systems may perform poorly when this context information is not covered by the training set or it is pronounced similarly to other terms.

Prior works to customize E2E ASR systems with contextual knowledge can be broadly classified into two categories. 
The first one is incorporating an external contextual language model (LM) into the E2E decoding framework to bias the recognition results towards context phrase list, which is generally implemented by adopting shallow fusion with a contextual finite state transducer (FST) \cite{OTF_rescore2, OTF_rescore3, CLAS, OTF_rescore4, le2021deep}. 
The second category is adding a context encoder which incorporates contextual information into E2E ASR systems \cite{CLAS, jain2020contextual, phoebe}. This method conducts contextual biasing in an E2E manner. However, it changes the source ASR model and \cite{CLAS} also reported scalability issues with large biasing phrase list.

Different from traditional methods, we propose to do contextual biasing on the ASR output with a contextual spelling correction model. We aim to make it an efficient, robust and general solution as a post processing ``plug-in'' module for E2E ASR customization.
To achieve this, an autoregressive (AR) and a non-autoregressive (NAR) contextual spelling correction model \cite{CSC} are proposed. The AR design, denoted as Contextual Spelling Correction (CSC), adds an additional context encoder into an AR E2E spelling correction model, which contains a text encoder, a context encoder, and a decoder. The contextual information is incorporated into the decoder by attending to the hidden representations from the context encoder with an attention mechanism \cite{Attention-bahdanau2014}. For the NAR model, denoted as Fast Contextual Spelling Correction (FCSC), we directly feed the output of text encoder into the decoder. The decoder attends to the context encoder and identifies locations that should be corrected and the candidate context index at each position. One of the advantages is that we can generate results in parallel and don’t need to conduct label-by-label prediction in AR, and the decoding speed is also increased. It should be noted that by FCSC we change this problem from a generation task to a classification task. CSC has the potential to correct any error made by the ASR system while FCSC focuses on biasing phrases, which is also a main difference between CSC and FCSC.
During inference, filter mechanisms for the context list are proposed to improve inference efficiency and deal with the scalability issues for large context lists. Performance balancing mechanisms are also proposed to adjust biasing degree and control possible WER regressions on general utterances that we do not want to bias.
In addition, we demonstrate that the model is a general contextual biasing solution which is effective among different domains.
Across several experiments, we find that the proposed method significantly outperforms the baseline contextual FST biasing method, and additional improvements can be achieved by further combining the proposed method on top of FST biasing, leading to the best performance. Additionally, compared to the AR model, the NAR solution reduces model size by 43.2\% and speeds up the inference by 2.1 times while achieving WER improvement.

\section{Related Work}

\subsection{Contextual LM Biasing}

A straightforward way to do contextual biasing is combining a contextual LM into the ASR system by shallow fusion, this language model is generally constructed by context phrases as an FST. For the E2E ASR system, it's implemented by interpolating the model posterior probabilities $P$ with the scores $P_c$ from an external contextual LM during beam-search decoding:
\begin{equation}
{y}^* = \mathop{\arg\max}_{y}{logP(y|x)}+\lambda{logP_{c}(y)},   
\end{equation}
where $\lambda$ is a tunable parameter which decides the weight of contextual LM. The contextual LM is constructed by compiling the list of biasing phrases into FST. This method is denoted as FST biasing below. 
In this research line, several techniques have been explored to improve the biasing performance for the E2E ASR system. \cite{OTF_rescore2} proposed to bias the E2E model by applying the contextual LM score boosts at word boundaries. This method cannot deal with proper nouns well because the E2E model uses graphemes or wordpieces during beam search and works at subword unit level.
To deal with this problem, \cite{CLAS} proposed to push the weights of the subword FST to each subword unit, which is found to be more effective in E2E ASR biasing. Techniques including adding failure arcs, biasing before beam pruning, biasing at the wordpiece level rather than grapheme level, and adding activation prefix to avoid regression on utterances that do not contain any biasing phrase (anti-context) are also explored to further improve the model performance \cite{CLAS, OTF_rescore4}.

\subsection{Bias Encoder}

Contextual biasing through shallow fusion is not jointly optimized with the training of the ASR model, which goes against the benefits of direct objective optimization of E2E models. To address this problem, the methods that incorporate contextual information by introducing an additional bias encoder into the E2E framework were proposed, such as CLAS \cite{CLAS}, contextual RNN-T \cite{jain2020contextual} and Phoebe \cite{phoebe}. This kind of method trains the E2E ASR model together with a bias encoder which encodes a context list created from the transcripts associated to the utterances in the training batch, and the decoder attends to both the audio encoder and bias encoder to bias output distribution during decoding. In inference, the context phrase list is incorporated as the input of bias encoder. However, this method is more expensive in training and inference \cite{CLAS} and changes the structure of raw ASR models. Some other works, such as trie-based deep biasing \cite{trie_deep_biasing}, two-step memory enhanced model \cite{instant_one_shot}, tree-constrained pointer generator component \cite{tree_constrained} are also investigated to further improve the model performance.

\subsection{Spelling Correction and deliberation}

The current research on spelling correction and deliberation approaches mainly try to improve the general ASR accuracy without contextual information.
Spelling correction models correct errors in ASR outputs in an E2E manner. 
In this research line, \cite{SC_1} first proposed an LSTM-based seq-to-seq model which was trained on synthesized speech generated from text-only data. The results showed reasonable improvements compared to simple LM rescoring methods. After that, more structures and training strategies have been proposed, such as a transformer-based model \cite{SC_2} for a Mandarin speech recognition task, models initialized from a pre-trained BERT model \cite{SC_3} or RoBERTa model \cite{SC_4}, and so on. 
Different from spelling correction model, deliberation model \cite{deliberation_1, deliberation_2, deliberation_3} attends to both first-pass decoding hypothesis and acoustic features to further improve recognition accuracy. 
One disadvantage of spelling correction or deliberation approaches is that it's hard to do streaming, and there is possible extra model size and latency impact for E2E ASR system due to additional decoding pass.
To mitigate the latency issue, \cite{SC_5} proposed an NAR spelling correction model FastCorrect which adds a length predictor to predict the target token number for each source token and generate target sequence input. Though the latency is largely reduced, there is still some regression compared to the AR model.

\section{Methodology}

\subsection{Autoregressive Contextual Spelling Correction}

\subsubsection{Model structure}

\begin{figure}[ht]
  \centering
  \includegraphics[width=0.85\linewidth]{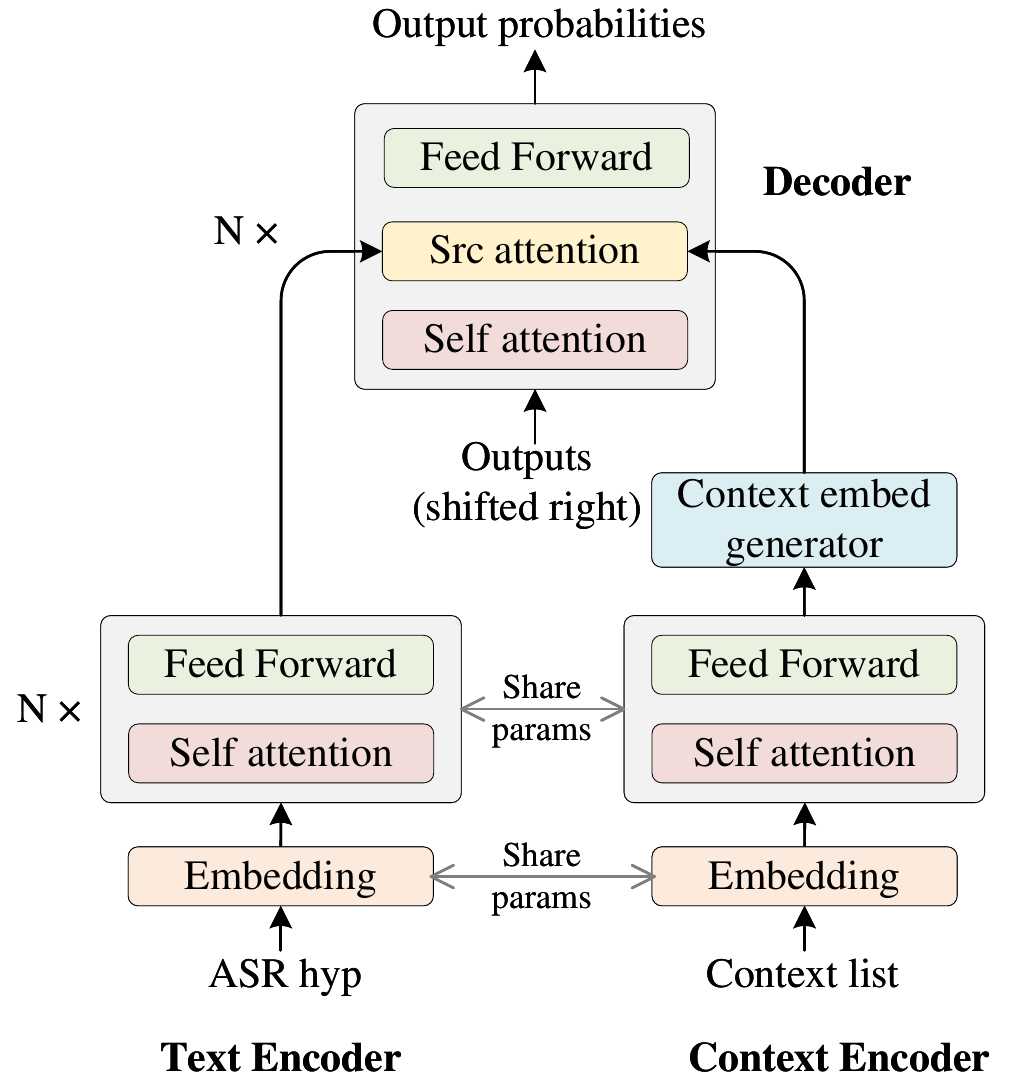}
  \caption{Autoregressive contextual spelling correction (CSC) model, which contains a text encoder, a context encoder and a decoder, the two encoders share parameters.}
  \label{fig:CSC_v1_structure}
\end{figure}

As shown in Figure \ref{fig:CSC_v1_structure}, this model is a seq2seq \cite{seq2seq} model with a text encoder, a context encoder, and a decoder, which takes the ASR hypothesis as the text encoder input and the context phrase list as the context encoder input. The context encoder encodes each context phrase as hidden states, these hidden states are averaged by context embedding generator to get context embedding. The decoder takes the output of the previous step as input autoregressively, and attends to the outputs of both encoders. These attentions are then added up to generate the final attention, from which the decoder obtains information from ASR hypothesis and context phrase embeddings to correct contextual misspelling errors. All the components are transformer-based \cite{attention} and composed by pre-LayerNorm \cite{layernorm, GPT2}, self-attention, encoder-decoder attention and feed-forward layer (FFN). Because the inputs of text encoder and context encoder are both transcriptions, it's natural to share the parameters of these two encoders, which reduces the model size and also helps context encoder training. The final loss is the cross entropy of output probabilities and ground truth label.

\subsection{Non-autoregressive Contextual Spelling Correction}

\subsubsection{Model structure}

\begin{figure}[ht]
  \centering
  \includegraphics[width=0.85\linewidth]{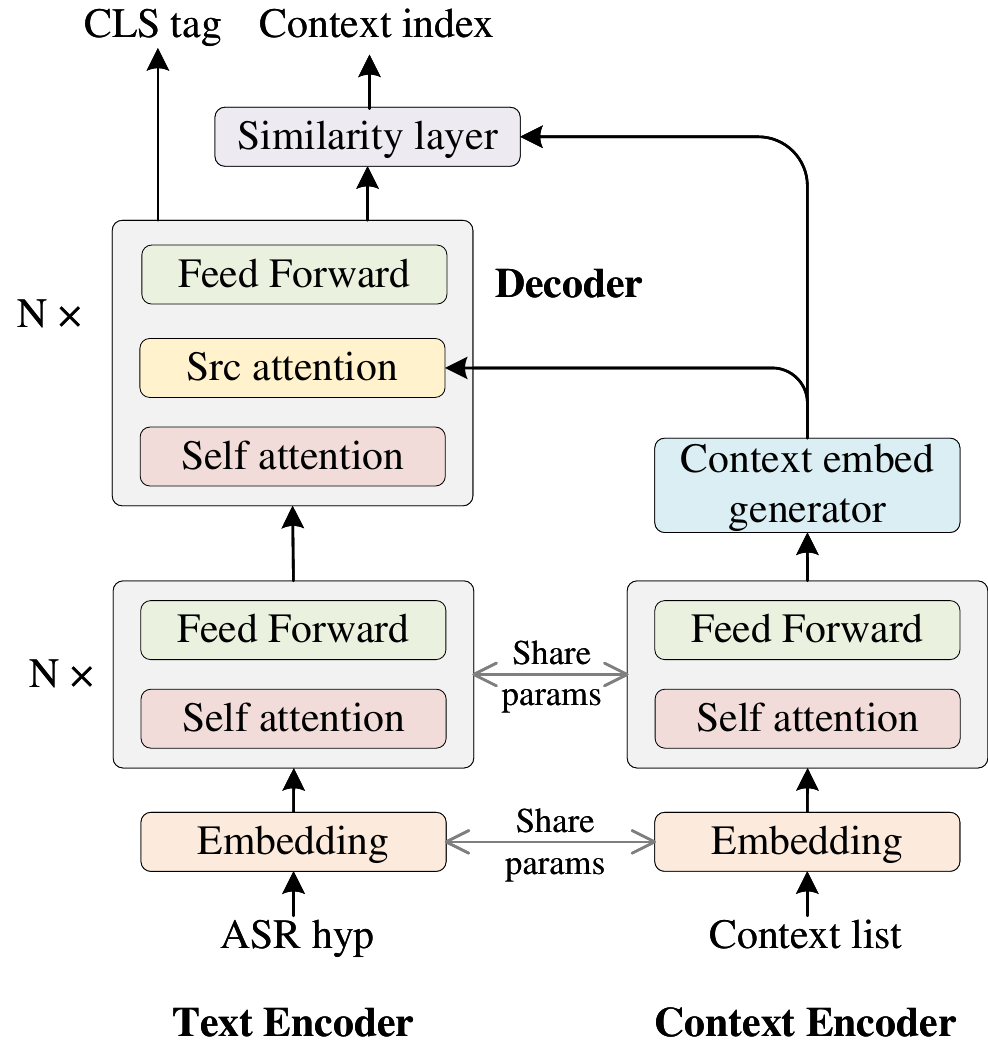}
  \caption{Non-autoregressive contextual spelling correction (FCSC) model, the decoder directly takes the text encoder output as input, and has two outputs: CLS tag and Context index.}
  \label{fig:CSC2}
\end{figure}

As shown in Figure \ref{fig:CSC2}, similar to AR CSC model, the proposed NAR model (FCSC) contains a text (ASR hypothesis) encoder, a context encoder and a decoder, where the text encoder takes ASR decoding results as input and the context encoder takes biasing phrase list as input. The parameters of the two encoders are shared. 
The decoder directly takes the output of text encoder as input and attends to the context encoder to decide where to correct and select which context phrase to correct. 
The encoder and decoder are both transformer-based, the output hidden states of each context phrase are averaged to obtain the context phrase embedding by context embedding generator. 
The similarity layer calculates the similarity of decoder output hidden states with context embeddings by an inner product operation:
\begin{equation}
s_{ij} = {\rm softmax}(\frac{Q_iW^Q(K_jW^K)^T}{\sqrt{d_k}}), \label{eqa:sim_layer}
\end{equation}
where $Q_i$ is the decoder hidden state at $i$-th position, $K_j$ is the $j$-th context phrase embedding, and $d_k$ is the dimension of $K$.

\subsubsection{Contextual biasing mechanism}

The decoder has two outputs: 

\textbf{CLS tag} $\boldsymbol{cls}$.
The position-wise classification (CLS) tag $cls$ has the same sequence length as input ASR hypothesis, which determines whether to correct the token at this position or not. It uses “BILO” representation where “B”, “I” and “L” represent the beginning, inside and last position of a context phrase, “O” represents a general position outside of a context phrase.

\textbf{Context index} $\boldsymbol{cind}$.
Context index is the output of similarity layer, which is the expected index of the ground-truth context phrase in the bias list for this position. We add an empty context at the beginning of the bias list, hence the context index for general tokens that should not be corrected is 0. As shown in Equation \ref{eqa:sim_layer}, the output hidden dimension of similarity layer at each position $i$ is the same as the input bias list size, and the context phrase corresponding to the largest value in $s_i$ is selected during decoding for this position:
\begin{equation}
cind_i = \mathop{\arg\max}{s_i}.
\end{equation}

According to the output CLS tag and context index, the final correction results can be determined by replacing the words tagged by CLS tag $cls$ with the context phrase selected by context index $cind$. Here is an input/output example:

\begin{table}[h]
  \label{tab:input_example}
  \centering
  \resizebox{\linewidth}{!}{%
  \begin{tabular}{ll}
    \toprule
    biasing phrases $c$ & \{\textit{“”, “Jack”, “\textcolor{blue}{Joe Biden}”, ..., “Tom Jones”}\}  \\
    ASR hyp $x$ & [   \textit{“\_who”}, \textit{“\_is”}, \textit{“\textcolor{blue}{\_john}”}, \textit{“\textcolor{blue}{\_b}”}, \textit{“\textcolor{blue}{ide}”}       ]  \\
    CLS tag $cls$ & [ \quad  O, \quad\quad O, \quad\quad\ \ B, \quad\quad I, \quad\quad L \  ] \\
    Context index $cind$ & [ \quad 0, \ \quad\quad 0, \quad\quad\ \ 2, \quad\quad 2, \quad\quad 2 \ \  ] \\
    Final output & [   \textit{“\_who”}, \textit{“\_is”}, \textit{“\textcolor{blue}{\_joe}”}, \textit{“\textcolor{blue}{\_b}”}, \textit{“\textcolor{blue}{id}”}, \textit{“\textcolor{blue}{en}”}  ] \\
    \bottomrule
  \end{tabular}}
\end{table}

In this example, the context phrase “\textit{Joe Biden}” is recognized as “\textit{John Bide}” by the ASR system, and the index of target context phrase “\textit{Joe Biden}” is 2 in the bias list. By design, the wordpieces of “\textit{John Bide}” should be labeled as [B,I,L] for CLS tag, and the corresponding context index output should be the same to where “\textit{Joe Biden}” is in context list. 
It should be noted that the final output sequence length can be different from the input due to the replace operation.

The final loss function is the sum of CLS tag loss and context index loss:
\begin{equation}
L = H(\widehat{y_{cls}}, y_{cls}) + H(\widehat{y_{cind}}, y_{cind}).
\end{equation}

\subsection{Data Processing and Training Strategy}
\label{sec:dataprocess}

For both CSC and FCSC, we first generate reference-hypotheses pairs from a large set of phrases by using a TTS system. Then the training data is generated during training by combining these context reference-hypotheses pairs with general scripts or sentence patterns. We also use teacher-student learning and quantization to reduce the model size and speed up inference.

\subsubsection{Text to speech}

To prepare the training data, we first collect a large number of context phrases, then a multi-speaker multi-locale (en-*, including en-us, en-gb, en-in, etc.) text to speech (TTS) system \cite{liu2021delightfultts} is adopted to generate TTS audio for these phrases. The generated TTS data are then fed into the ASR model to get recognition hypotheses. The input context phrase and the generated hypotheses are paired to get context reference-hypotheses pairs, examples like “\textit{John}”--[“\textit{Jane}”, “\textit{Jon}”, “\textit{June}”].
It should be noted that most context phrases are much shorter than training scripts, which leads to a fast data preparation process both in TTS data generation and ASR inference.

\subsubsection{Training pairs construction}

We also prepare a set of sentence patterns and general scripts to generate training scripts together with the prepared context reference-hypotheses pairs during training. For sentence patterns, we fit the context phrase and one of its hypotheses into the pattern. For general scripts, we randomly replace words with context phrase and its hypothesis. Here is an example: 

\begin{table}[h]
  \label{tab:training_pair_example}
  \centering
  \resizebox{0.9\linewidth}{!}{%
  \begin{tabular}{ll}
    \toprule
    Context ref-hyp pair & “\textit{John}” -- [“\textit{Jane}”, “\textit{Jon}”, “\textit{June}”]  \\
    \midrule
    Pattern & \textit{Call \textcolor{red}{$<$PersonName$>$} at ten a.m.}  \\
    Hyp $x$ & \textit{Call \textcolor{blue}{Jon} at ten a.m.} \\
    Ref $y$ & \textit{Call \textcolor{blue}{John} at ten a.m.} \\
    \midrule
    General script & \textit{When do you come to \textcolor{red}{me}}  \\
    Hyp $x$ & \textit{When do you come to \textcolor{blue}{Jane}} \\
    Ref $y$ & \textit{When do you come to \textcolor{blue}{John}} \\
    \bottomrule
  \end{tabular}}
\end{table}

Where “\textit{Jon}” and “\textit{Jane}” are randomly selected hypothesis of context phrase “\textit{John}” for the sentence pattern and general script, “\textit{me}” is the randomly selected word to be replaced for the general script.

To take care of utterances that do not contain any context phrase but with a biasing phrase list (anti-context) and cases that the target context phrase somehow doesn't appear in the biasing phrase list, we also leave part of the input general scripts in the training set unchanged with probability $P_{cont}$. 

\subsubsection{Context setup}

During training, we randomly sample $N_c$ biasing phrases for each utterance from the whole context list which is pooled from biasing phrases of all utterances in this batch. $N_c$ is randomly sampled from uniform distribution $[1, N_{cmax}]$, where $N_{cmax}$ is the max context list size defined as a training hyper-parameter. To distinguish the sampled context list for each utterance in the batch, a context mask is adopted on top of the context embedding generator. 
What's more, to increase the diversity of possible error patterns, besides using multi-speaker multi-locale TTS system to generate TTS data, we also swap the reference context phrase and its hypothesis randomly with probability $P_m$ to extend the set of context reference-hypotheses pairs.

\subsubsection{Teacher-student learning}

Teacher-student learning \cite{li2014learning, hinton2015distilling} is an effective way to reduce model size and improve inference efficiency. For both CSC and FCSC, we use teacher-student learning to make it smaller and more efficient. The loss function to train the student model contains a hard loss $L_{hard}$ and a soft loss $L_{soft}$. The hard loss is the cross-entropy of student model output $y_{S}$ and reference $y$, and the soft loss is the KL-divergence of $y_{S}$ and teacher model output $y_{T}$:
\begin{equation}
L = {\alpha}L_{soft}+(1-\alpha)L_{hard}   
\end{equation}
\begin{equation}
L_{hard} = H(y_{S}, y)   
\end{equation}
\begin{equation}
L_{soft} = D_{\rm KL}\left({\rm softmax}(\frac{y_{S}}{T}), {\rm softmax}(\frac{y_{T}}{T})\right) \cdot T^2
\end{equation}

where $T$ is the temperature to adjust the smoothness of output probabilities, $\alpha$ determines the proportion of hard loss and soft loss. For FCSC, the final hard/soft loss is the sum of the loss of CLS tag $cls$ and Context index $cind$:
\begin{equation}
L_{hard} = L_{hard}^{cls}+L_{hard}^{cind}
\end{equation}
\begin{equation}
L_{soft} = L_{soft}^{cls}+L_{soft}^{cind}
\end{equation}

\subsection{Inference}

\subsubsection{Context pre-selection mechanism}

For context phrase list, we propose a relevance ranker (rRanker) and a preference ranker (pRanker) to preselect context phrases from the raw context list size and deal with the possible scalability issue. The preference ranker represents the preliminary knowledge of the context phrases in terms of preference weight, examples like the call frequency of context phrases in the target domain. The relevance ranker aims to measure the relevance between the specific ASR hypothesis and the given context phrases. To simplify the inference process and control latency, we propose an edit distance-based method to obtain the weight of relevance ranker:
\begin{equation}
W_r^j = -\frac{\min_{i}({\rm edit\_distance}(c_j, e_i))}{{\rm len}(c_j)},
\end{equation}
where $e_i$ is the segment cut off from input ASR hypothesis with the same length of the context phrase $c_j$ from the $i$-th word. The final relevance ranker weight is the minimum value of these edit distance normalized by the length of $c_j$. For edge cases where the length of remaining characters from the $i$-th word is shorter than $c_j$, we simply use the remaining characters as $e_i$. Here is an example: 
\vspace{2pt}

\quad \quad \quad \quad \textcolor{red}{$e_1$}  \quad \quad \quad \quad \textcolor{red}{$e_3$}  \quad \quad \quad \quad \textcolor{red}{$e_6$}

\quad \quad \quad \quad \textcolor{blue}{\underline{Please }}send \textcolor{blue}{\underline{a messa}}ge to \textcolor{blue}{\underline{Ernest}}

\quad \quad \quad \quad \textcolor{red}{$c_j$}:\ \textcolor{blue}{Earnest}

\vspace{2pt}
where $e_1$ and $e_3$ are the first and third segments cut off from the first and third word with the same length of $c_j$. $e_6$ is an edge case which is shorter than $c_j$. The minimum edit distance should be obtained at $e_6$. After that, the preference ranker weight $W_p$ and relevance ranker weight $W_r$ are combined with weight ${\alpha}_p$ and $1-{\alpha}_p$, respectively. The top $K_f$ context phrases are then selected out from the raw $K_r$ bias phrases as the final input for context encoder:
\begin{equation}
c_1, c_2, ..., c_{K_f} = \mathop{\rm \arg topK}\limits_{j}(\alpha_{p}W_{p}^{j}+(1-\alpha_{p})W_{r}^{j}).
\end{equation}

\subsubsection{FCSC output format}

For FCSC, the final output is determined based on the model outputs CLS tag $cls$ and context index $cind$. However, during inference, the output format is not always as standard as what training set looks like, which is referred to as illegal output. Some examples are listed in Table \ref{tab:output_example}. This happens when the model is not that confident of the output, so we will give up correcting such cases.

\begin{table}[h]
  \caption{output examples of $cls$ and $cind$ for fcsc}
  \label{tab:output_example}
  \centering
  \resizebox{\linewidth}{!}{%
  \begin{tabular}{lll}
    \toprule
    Outputs & Comments & Legal or not \\
    \midrule
    $cls$:\quad [O, O, B, I, L, O] & correct & \Checkmark \\
    $cind$: $[0, 0, 9, 9, 9, 0]$ & & \\
    \midrule
    $cls$:\quad [O, O, B, I, I, O] & incomplete $cls$ where the & \XSolid \\
    $cind$: $[0, 0, 6, 6, 6, 0]$ & end position tag L is lost & \\
    \midrule
    $cls$:\quad [O, O, B, L, O] & $cls$ and $cind$ inconsistency & \XSolid \\
    $cind$: $[0, 4, 4, 0, 0]$ &  & \\
    \midrule
    $cls$:\quad [O, B, I, L] & tagged position corresponds & \\
    $cind$: $[0, 4, 5, 4]$ & to multiple context indexes & \XSolid \\
    \bottomrule
  \end{tabular}}
\end{table}

\subsubsection{Performance balancing}

When the model is too “biased”, the model may suffer from regressions on anti-context cases. FST biasing generally uses an interpolation weight $\lambda$ to balance the model performance among biasing sets and anti-context cases. For the proposed method, a part of training set are general scripts without context phrase, which teaches the model to decide when to correct according to the given context phrase list and the input ASR hypothesis, hence the model can effectively control the regression on anti-context cases by itself. However, regressions still exist for some cases and there may be additional requirements on the regression in real application. Here we propose to use NER detector and controllable parameter $s^o$ to control the regression of CSC/FCSC. We will show the effectiveness of these methods in the experiment section below.

\textbf{NER detector}.

The first solution is adopting an NER detector before CSC/FCSC inference, which classifies whether the ASR hypothesis contains terms to be biased and decides when to activate CSC/FCSC.
In general, this detector can be designed as a classification model, specifically, a Named Entity Recognition (NER) task \cite{NER1, NER2} which classifies entities in the given ASR hypothesis. The training data for this classification model can be obtained following the methods in NER tasks.
For narrow domains with representative sentence patterns, rule-based detector is enough to achieve reasonable classification precision and recall. The ``bias prefixes'' used by \cite{OTF_rescore4,CLAS} is indeed a subset of rule-based detector. 
However, just prefix information is limited, especially for cases that without prefix (e.g., ``\textit{\textcolor{blue}{Harry} is my good friend}'') and cases that with too general prefix (e.g., ``\textit{schedule a meeting for me and \textcolor{blue}{Jessa} on Tuesday}'' where ``\textit{and}'' is the prefix).
For the proposed model, we can use prefixes, suffixes, and sentence patterns to make a more reliable classification because we can see the full ASR hypothesis. 
In our experiments, some of the baseline ASR models are able to tag out entities with classification tokens such as $<$\textit{name}$>$ and $<$\textit{entity}$>$, which can be directly adopted as NER detector, denoted as \textbf{token filter (tfilter)}. For baseline ASR models without such tokens, a rule-based detector is used for convenience, denoted as \textbf{pattern filter (pfilter)}. What's more, for multiple output candidates of an ASR system, CSC/FCSC is activated as long as one of these candidates passes through the detector.

\textbf{Controllable threshold} $\boldsymbol{s^o}$.

Although the NER detector has effective regression control ability for anti-context cases, it's still time consuming and not easy to extract the prefixes/suffixes/patterns from an outside adaptation set, there are also many things to consider to balance the classification precision and recall. Therefore, for FCSC, we also propose another mechanism to easily control the regression by just adjusting a controllable threshold parameter $s^o$, which measures the confidence of the model output by adopting the output probabilities $s_{ij}$ from similarity layer:

\begin{equation}
s_i^* = \mathop{\rm mean}\limits_{i}(\max_j(s_{ij})), ~~ i \in [i_s, i_e]
\end{equation}

\begin{equation}
cind_i = \left\{ 
\begin{array}{rcl}
\mathop{\arg\max}{s_i} & & {s_i^*>=s^o} \\
0 & & {s_i^*<s^o} \\
\end{array} \right.
\end{equation}

Where $[i_s,i_e]$ is a continuous sequence labeled out by CLS tag, for example, for output \{$cls$: $[O,O,B,I,L,O]$\}, $i_s$ is 2 and $i_e$ is 4. $s_i^*$ represents the confidence of the model for the current output, and $s^o$ is a threshold value which ranges from 0 to 1. This threshold can be tuned for each application scenario to meet the regression requirements. When $s_i^*<s^o$, this position is left unchanged due to low confidence.

\subsubsection{Score interpolation}

ASR system may output multiple candidates for each utterance, these candidates contain more signals and help improve the model performance according to our experiments. 
However, too many input candidates will increase the latency, we select top $N_{asr}$ candidates for CSC/FCSC decoding, where $N_{asr}$ is determined by the model performance and latency requirements. 

CSC conducts beam search process during decoding, which generates the corresponding CSC hypotheses $\{H_{i1}, H_{i2}, ..., H_{iN_{csc}}\}$ for each ASR hypothesis $H_i$. The final decoding results are obtained by ranking these $N_{asr}\times{N_{csc}}$ hypotheses:
\begin{equation}
H^* = \mathop{\arg\max}_{H}{\lambda_{ASR}{logp_i}+\lambda_{CSC}{log{p'}_{ij}}}
\end{equation}
where $\lambda_{ASR}$ and $\lambda_{CSC}$ are the weights for ASR and CSC scores. $p_i$ is the utterance level probability of $H_i$, and ${p'}_{ij}$ is the utterance level probability of the $j$-th CSC hypothesis for $H_i$.

Unlike the AR method, there is no beam search process for FCSC. 
The most possible hypothesis is obtained by the following criteria:

\begin{equation}
H^* = \mathop{\arg\max}_{H}{\lambda_{ASR}{logp_i}+\lambda_{FCSC}{logq_{i}}}
\end{equation}
where $logq_i$ is the utterance level FCSC score which is the sum of utterance level log probabilities of $cls$ and $cind$:

\begin{equation}
logq_i = logq_i^{cls} + logq_i^{cind}
\end{equation}

\section{Experiment}

\subsection{Data Sets}

The training context list includes 1M names generated from 48k name words, and 6M proper nouns collected from two open source datasets NELL \cite{NELL} and Yago \cite{Yago}. We use a multi-speaker multi-locale(en-*) TTS \cite{liu2021delightfultts} AM which contains 363 speakers to get TTS audio, the generated TTS data are then fed into an RNN-T model to obtain ASR hypotheses. A context phrase dictionary is then constructed from these reference-hypotheses pairs, in which each reference context phrase corresponds to a list of hypotheses. What's more, 512 sentence patterns and 26M in-house general scripts are used to construct training set with the generated context phrase dictionary as the method mentioned before. We have open-sourced the TTS synthetic data for the collected entities\footnote{\url{https://github.com/zombbie/entity-synthetic-dataset}}. 

We evaluate model performance on several test sets, detailed in Table \ref{tab:test_sets}. The Random8k set contains anti-context utterances, the context list of this set is randomly selected from Name set, which aims to ensure that the model doesn't affect recognition quality if all biasing phrases are irrelevant. To evaluate the model performance on general context biasing in other domains, there are also two test sets in which biasing phrases are general phrases rather than person names or proper nouns, the biasing phrases of these sets are frequently used terms in its own domain but not related to training set. 

\begin{table*}[ht]
  \caption{test sets}
  \label{tab:test_sets}
  \centering
  \begin{tabular}{lllll}
    \toprule
    Test set & \#utt. & context type & context list size & context examples  \\
    \midrule
    Name set & $11597$ & person name & $1509$ &  Jotham Parker  \\
    Personal assistant (PA) dev & $830$ & person name & $1507$ & Andrew \\
    Personal assistant (PA) blind & $1024$ & person name & $1507$ & Xiaofang \\
    Random8k set & $8000$ & - & - & - \\
    Text editor & $797$ & commands & $210$ & Uppercase the next paragraph \\
    Medical set & $516$ & medical health related & $11638$ & Pause cardiac output monitoring \\
    \bottomrule
  \end{tabular}
\end{table*}

\subsection{Experimental Setup}

During training, we set the batch size to 300, the max size of context phrase list $N_{cmax}$ of each utterance to 100, the ratio of general utterances without context phrase $P_{cont}$ to 20\%, and the ratio of swapping context reference-hypotheses pairs $P_m$ to 20\%. We use teacher-student learning and quantization to reduce model size. The student model was trained with the same training set as teacher, and we set $T$ to 1 and $\alpha$ to 0.9. The model parameters of teacher and student are listed in Table \ref{tab:model_parameters}. 

During inference, we set the size of filtered context list $K_f$ to 100 to be consistent with training. We decode with two baseline ASR models: an RNN-T model and a T-T model, trained with 64 thousand hours Microsoft anonymized data. The RNN-T model is able to output $<$\textit{name}$>$ tokens which classifies person names (tfilter), which can be directly used as NER detector for both CSC and FCSC. A pattern filter (pfilter) which is generated from an outside adaptation set is adopted for T-T model. Similarly, the baseline FST biasing method uses $<$\textit{name}$>$ tokens for RNN-T and prefix set extracted from the same adaptation set for T-T as activation prefix. For RNN-T, top 4 candidates are selected to correct, while for T-T, we select top 1 because the T-T model we use just supports single candidate output. We also test FCSC model performance when with the proposed controllable threshold $s^o$ rather than NER detector.

\begin{table}[h]
  \caption{model parameters}
  \label{tab:model_parameters}
  \centering
  \resizebox{\linewidth}{!}{%
  \begin{tabular}{ccccccc}
    \toprule
    Model & & layers & d\_model & heads & d\_FFN & Params(M)  \\
    \midrule
    \multirow{2}{*}{CSC} & Teacher & $6$ & $512$ & $8$ & $2048$ & $55.4$  \\
    & Student & $3$ & $192$ & $4$ & $768$ & $5.2$ \\
    \midrule
    \multirow{2}{*}{FCSC} & Teacher & $6$ & $512$ & $8$ & $2048$ & $47.4$  \\
    & Student & $3$ & $192$ & $4$ & $768$ & $4.2$ \\
    \bottomrule
  \end{tabular}}
\end{table}

\subsection{Performance}

\subsubsection{WER reduction}

The results on the first three test sets in name domain and the Random8k set are listed in Table \ref{tab:perf_name}, where ``Bias'' represents the baseline FST biasing method. Compared to baseline ASR models, the proposed model achieves as much as 41.7\% relative word error rate reduction on RNN-T and 51.0\% on T-T. Compared to the FST biasing method, the proposed model achieves comparable or much better performance among all these sets. We also test the model performance based on RNN-T+Bias and T-T+Bias, and find our method can still achieve significant gain based on FST biasing as much as 32.6\% on RNN-T and 38.5\% on T-T, which indicates that the proposed model can work on top of the FST biasing method to achieve the best performance for E2E ASR customization. Compared to AR CSC model, the NAR solution still performs better among all the test sets, which is slightly different from other NAR solutions. The reason, we believe, is that different from other NAR architectures, our method doesn't need to align the encoder and decoder by adopting a length or duration predictor, which avoids possible error accumulation, the structure we use is also very suitable for this task.

Table \ref{tab:perf_other} lists the results on the two general biasing sets. Compared to baseline RNN-T, the model achieves as much as 50.7\% relative WER improvement. When correct on top of FST biasing, 24.6\% relative WER improvement can also be achieved. These results demonstrate that the proposed method is a general contextual biasing solution which is domain-insensitive and is effective among different scenarios. 

\begin{table*}[htb]
  \caption{model performance on name domain}
  \label{tab:perf_name}
  \centering
  \begin{tabular}{lllllllll}
    \toprule
    \multirow{2}{*}{Method}  & \multicolumn{2}{c}{Name set}  & \multicolumn{2}{c}{PA dev} & \multicolumn{2}{c}{PA blind} & \multicolumn{2}{c}{Random8k set}   \\
    \cmidrule(r){2-3}\cmidrule(r){4-5}\cmidrule(r){6-7}\cmidrule(r){8-9}
     & \multicolumn{1}{c}{WER} & \multicolumn{1}{c}{WERR} & \multicolumn{1}{c}{WER} & \multicolumn{1}{c}{WERR} & \multicolumn{1}{c}{WER} & \multicolumn{1}{c}{WERR} & \multicolumn{1}{c}{WER} & \multicolumn{1}{c}{WERR}   \\
    \midrule
    RNN-T      & $30.2$ & - & $24.2$ & - & $22.5$ & - & $13.9$ & -   \\
    \quad +CSC  & $18.1$ & $40.1$ & $17.4$ & $28.1$ & $15.3$ & $32.0$ & $14.4$ & $-3.6$     \\
    \quad +FCSC  & $17.6$ & $41.7$ & $16.8$ & $30.6$ & $14.6$ & $35.1$ & $14.4$ & $-3.6$     \\
    \quad +FCSC ($s^o$)  & $18.0$ & $40.4$ & $17.3$ & $28.5$ & $15.2$ & $32.4$ & $14.0$ & $-0.7$     \\
    RNN-T+Bias  & $23.9$ & $20.8$ & $19.5$ & $19.4$ & $17.1$ & $24.0$ & $14.0$ & $-0.7$     \\
    \quad +CSC  & $16.9$ & $44.0$ & $16.7$ & $31.0$ & $14.1$ & $37.3$ & $14.6$ & $-5.0$     \\
    \quad +FCSC  & $16.1$ & $46.7$ & $15.5$ & $36.0$ & $13.1$ & $41.8$ & $14.4$ & $-3.6$     \\
    \quad +FCSC ($s^o$)  & $16.3$ & $46.0$ & $15.9$ & $34.3$ & $13.1$ & $41.8$ & $14.1$ & $-1.4$     \\
    \midrule
    T-T      & $24.1$ & - & $17.5$ & - & $16.6$ & - & $8.8$ & -   \\
    \quad +FCSC & $11.8$ & $51.0$ & $12.7$ & $27.4$ & $11.2$ & $32.5$ & $8.8$ & $-0.0$     \\
    \quad +FCSC ($s^o$) & $11.9$ & $50.6$ & $11.9$ & $32.0$ & $11.3$ & $31.9$ & $9.0$ & $-2.3$     \\
    T-T+Bias  & $17.4$ & $27.9$ & $12.6$ & $27.9$ & $12.6$ & $24.0$ & $8.9$ & $-1.1$     \\
    \quad +FCSC & $10.7$ & $55.6$ & $10.4$ & $40.6$ & $10.3$ & $38.0$ & $9.0$ & $-2.3$     \\
    \quad +FCSC ($s^o$) & $10.7$ & $55.6$ & $9.9$ & $43.4$ & $10.2$ & $38.6$ & $9.0$ & $-2.3$     \\
    \bottomrule
  \end{tabular}
\end{table*}

\begin{table}[t]
  \caption{model performance on general biasing sets}
  \label{tab:perf_other}
  \centering
  \begin{tabular}{lllll}
    \toprule
    \multirow{2}{*}{Method}  & \multicolumn{2}{c}{Text editor}  & \multicolumn{2}{c}{Medical set}   \\
    \cmidrule(r){2-3}\cmidrule(r){4-5}
    & \multicolumn{1}{c}{WER} & \multicolumn{1}{c}{WERR} & \multicolumn{1}{c}{WER} & \multicolumn{1}{c}{WERR} \\
    \midrule
    RNN-T      & $14.6$ & - & $14.0$ & -  \\
    \quad +FCSC  & $7.2$ & $50.7$ & $8.6$ & $38.6$     \\
    RNN-T+Bias  & $10.3$ & $29.3$ & $6.9$ & $50.8$     \\
    \quad +FCSC  & $7.9$ & $45.9$ & $5.2$ & $62.9$     \\
    \bottomrule
  \end{tabular}
\end{table}

\subsubsection{Latency improvement}

The model size and latency of the proposed AR CSC model and the NAR FCSC model are listed in Table \ref{tab:perf_runtime}. The latency is the decoding time of CSC/FCSC model per utterance in the test set regardless of baseline ASR model, which is tested on a machine with 2.60GHz CPU using single thread, each utterance corresponds to 4 ASR output candidates. The NAR model reduces model size by 43.2\% and speeds up the inference by 2.1 times, which shows huge advantages for the NAR model to be deployed on device compared to the AR solution. It should be noted that this latency includes context phrase encoding, which can be calculated in advance and loaded as cache in real application, therefore, the latency can be further reduced and optimized in runtime.

\begin{table}[t]
  \caption{model size and latency}
  \label{tab:perf_runtime}
  \centering
  \begin{tabular}{llll}
    \toprule
     & model & teacher  & student   \\
    \midrule
    \multirow{2}{*}{size(MB)} & CSC & $223.0$ & $9.5$   \\
     & FCSC & $212.0$ & $5.4$ \\
    \midrule
    \multirow{2}{*}{latency(ms/utt)} & CSC & $793.6$ & $132.3$   \\
      & FCSC & $530.0$ & $63.7$     \\
    \bottomrule
  \end{tabular}
\end{table}

\subsubsection{Influence of context list size}

\begin{figure}[h]
  \centering
  \includegraphics[width=\linewidth]{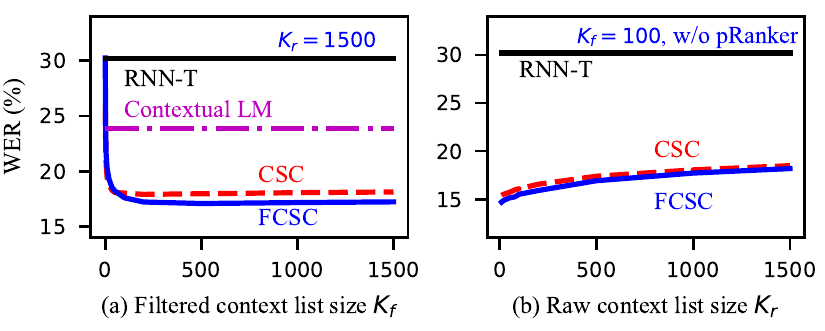}
  \caption{Influence of filtered context list size $K_f$ and raw context list size $K_r$ on moder performance. WER increases with Kr in a concave curve shape, and FCSC achieves stable improvement over CSC.}
  \label{fig:contsize}
\end{figure}

Figure \ref{fig:contsize} illustrates the influence of filtered context list size $K_f$ and raw context list size $K_r$ on Name set. Where $K_r$ depends on the given context phrase list, and $K_f$ is a decoding parameter that can be directly adjusted during inference. Here $K_r$ is adjusted by randomly sampling a subset of context phrases from the raw context phrase list while keeping the target phrase in the set. We can see WER increases with $K_r$ in a concave curve shape for both CSC and FCSC, which means raw context list size does not influence the model performance much when it's large enough, the proposed method can deal with the scalability problem well. The figure also shows the WER curve of FCSC lies below CSC in most of time for both $K_r$ and $K_f$, which indicates FCSC achieves stable improvement over CSC with decoding parameter change.

\subsubsection{Performance balancing of FCSC}

\begin{figure}[h]
  \centering
  \includegraphics[width=\linewidth]{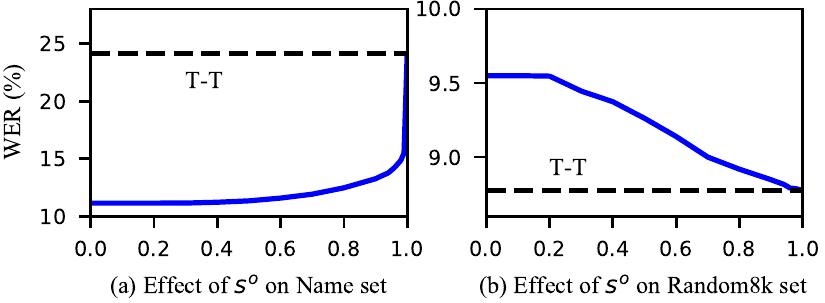}
  \caption{Control model performance among different test sets with threshold $s^o$. With the increase of $s^o$, the WER increases slowly on Name set while decreases quickly on Random8k set (which contains anti-context cases).}
  \label{fig:control}
\end{figure}

\begin{figure}[h]
  \centering
  \includegraphics[width=0.55\linewidth]{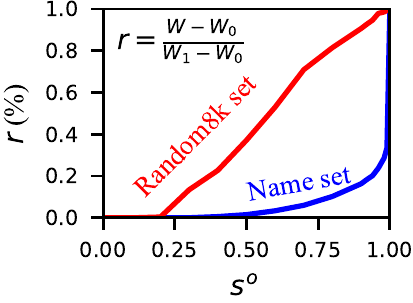}
  \caption{WER relative change with $s^o$. The WER on Name set experiences a long flat period and then encounters steep increase only when near to 1.0, which guarantees enough safe space to conduct performance balancing.}
  \label{fig:control_ratio}
\end{figure}

As discussed before, for FCSC, we propose a regression control mechanism which adopts controllable threshold $s^o$ to balance model performance between biasing set and anti-context cases. The model performance with $s^o=0.7$ on name domain is listed in Table \ref{tab:perf_name}, which shows similar performance with the method that adds NER detector. Figure \ref{fig:control} illustrates the effect of $s^o$ to model performance on Name set and Random8k set in details. $s^o=0$ is the vanilla setting which means we don't use this mechanism while $s^o=1$ means we totally give up correction. With the increase of $s^o$, the WER increases on Name set while decreases quickly on Random8k set, which demonstrates the effectiveness of the proposed balance mechanism. We define $r$ as the relative WER gap narrowing ratio:
\begin{equation}
r = \frac{W-W_0}{W_1-W_0},
\end{equation}
where $W$ represents WER, $W_0$ and $W_1$ represent the WER when $s^o=0$ and $s^o=1$. The variation of $r$ with the change of $s^o$ is illustrated in Figure \ref{fig:control_ratio}. We can see the WER on Name set experiences a long flat period when $s^o$ is small and then encounters steep increase only when near to $1.0$, which means most of the cases in Name set are corrected with enough confidence and guarantees enough safe space for us to conduct performance balancing.  

\subsubsection{OOV performance}

Table \ref{tab:perf_oov} lists model performance of out-of-vocabulary (OOV) terms and in-vocabulary (IV) terms on different test sets, where OOV rate represents the proportion of utterances with OOV biasing phrases not seen in CSC/FCSC model training. It should be noted that most of the bias phrases in Text editor set and Medical set are not seen in training set. 
The results show good performance on OOV terms, which indicates that the model learns error patterns in wordpiece level rather than word level. 
We also find that the performance on IV terms is generally worse, we checked the results and the explanation is that for the three sets with names, the raw WER of IV terms is relatively low and contains less error for correction; for the Text editor set and Medical set, there are only few samples in IV set and not that representative.
In general, OOV terms will easily appear in domains which are not related to training, the OOV performance also demonstrates the general contextual biasing ability of the proposed method from another point of view.

\begin{table}[h]
  \caption{oov performance (wer)}
  \label{tab:perf_oov}
  \centering
  \begin{tabular}{lllll}
    \toprule
    dataset  & OOV rate & model & OOV  & IV   \\
    \midrule
    \multirow{2}{*}{Name set} & \multirow{2}{*}{$54.0\%$} & T-T & $32.7$ & $16.8$   \\
      &  & +FCSC & $15.3$ & $11.3$     \\
    \midrule
    \multirow{2}{*}{PA dev} & \multirow{2}{*}{$42.4\%$} & T-T & $24.5$ & $13.3$   \\
      &  & +FCSC & $16.2$ & $11.0$     \\
    \midrule
    \multirow{2}{*}{PA blind} & \multirow{2}{*}{$32.3\%$} & T-T & $26.8$ & $11.1$   \\
      &  & +FCSC & $15.6$ & $9.5$     \\
    \midrule
    \multirow{2}{*}{Text editor} & \multirow{2}{*}{$82.6\%$} & RNN-T & $14.7$ & $14.1$   \\
      & & +FCSC & $6.4$ & $11.6$     \\
    \midrule
    \multirow{2}{*}{Medical set} & \multirow{2}{*}{$93.8\%$} & RNN-T & $14.1$ & $10.7$   \\
      & & +FCSC & $8.6$ & $8.9$     \\
    \bottomrule
  \end{tabular}
\end{table}

\section{Conclusions}

In this paper, we introduce a novel contextual biasing method for customizing end-to-end ASR systems. Our method integrates a context encoder into the spelling correction model with carefully designed AR and NAR mechanisms. Novel filtering algorithms are designed to deal with the large size context list. Effective performance balancing mechanisms are proposed to balance model performance on anti-context terms. The method is also a general contextual biasing solution which is domain-insensitive and can be adopted in different scenarios. Empirical studies have demonstrated that the proposed method significantly improves the ASR model performance on contextual biasing tasks, which also shows competitive or better performance over conventional methods. The NAR architecture achieves smaller model size, lower latency, and better performance than the AR design, which is more competitive for on-device application. 
However, there are still some limitations of the proposed method, including not being acoustic-grounded and the extra complexity in being streaming for the NAR solution. In future work, we would like to explore integrating additional acoustic information into the model to further improve the performance. We would also explore methods to enable streaming for the NAR model to further reduce latency.

\ifCLASSOPTIONcaptionsoff
  \newpage
\fi

\bibliographystyle{IEEEtran}

\bibliography{mybib}




\end{document}